\documentclass[10pt,twocolumn,letterpaper]{article}

\usepackage{wacv}
\usepackage{times}
\usepackage{epsfig}
\usepackage{graphicx}
\usepackage{amsmath}
\usepackage{amssymb}
\usepackage{url}

\usepackage{algorithm}
\usepackage[noend]{algpseudocode}

\wacvfinalcopy 

\def\wacvPaperID{114} 

\ifwacvfinal\fi
\begin{document}

\title{A Proposal-Based Solution to Spatio-Temporal \\ Action Detection in Untrimmed Videos }

\author{Joshua Gleason\textsuperscript{*},~~Rajeev Ranjan\textsuperscript{*},~~Steven Schwarcz\textsuperscript{*},~~Carlos D. Castillo,~~Jun-Cheng Chen, \\
Rama Chellappa \\
University of Maryland, College Park\\
{\tt\small gleason@umiacs.umd.edu, rranjan1@umiacs.umd.edu, schwarcz@umiacs.umd.edu}\\
{\footnotesize \textsuperscript{*}these authors contributed equally to this work}
}

\maketitle
\ifwacvfinal\thispagestyle{empty}\fi

\begin{abstract}

Existing approaches for spatio-temporal action detection in videos are limited by the spatial extent and temporal duration of the actions. In this paper, we present a modular system for spatio-temporal action detection in untrimmed security videos. We propose a two stage approach. The first stage generates dense spatio-temporal proposals using hierarchical clustering and temporal jittering techniques on frame-wise object detections. The second stage is a Temporal Refinement I3D (TRI-3D) network that performs action classification and temporal refinement on the generated proposals. The object detection-based proposal generation step helps in detecting actions occurring in a small spatial region of a video frame, while temporal jittering and refinement helps in detecting actions of variable lengths. Experimental results on the spatio-temporal action detection dataset - DIVA - show the effectiveness of our system. For comparison, the performance of our system is also evaluated on the THUMOS'14 temporal action detection dataset.

\end{abstract}

\vspace{-5mm}
\section{Introduction}
\vspace{-2mm}
\label{sec:intro}

Action detection in untrimmed videos is a challenging problem. Although methods using deep convolutional neural networks (CNNs) have significantly improved performance on action classification, they still struggle to achieve precise spatio-temporal action localization in challenging security videos. There are some major challenges associated with action detection from untrimmed security videos. First, the action typically occurs in a small spatial region relative to the entire video frame. This makes it difficult to detect the actors/objects involved in the action. Second, the duration of the action may vary significantly, ranging from a couple of seconds to a few minutes.
\begin{figure}[htp!]
    \centering
    \includegraphics[width=1\linewidth]{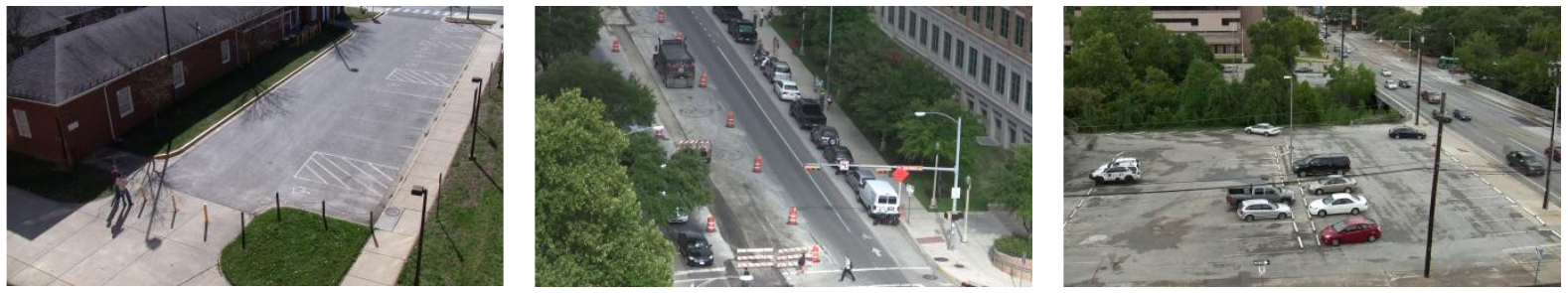}
    \caption{Sample images of different scenes of the DIVA dataset which presents challenging action detection scenarios which require algorithms to be robust to large variations in scale, object pose, and camera viewpoint.} 
\label{fig:sample_scene}
\vspace{-5mm}
\end{figure}
This requires the detection procedure to be robust to temporal variation. Existing publicly available action detection datasets such as THUMOS'14~\cite{thumos} and AVA~\cite{ava} do not posses these challenges. Hence, algorithms trained on these datasets have sub-optimal performance on untrimmed security videos. In this paper, we work with the DIVA dataset that has untrimmed security videos. Videos that comprise the DIVA dataset are a subset of those in the VIRAT dataset \cite{virat}, albeit with newly introduced annotations that make them suitable for the activity detection task. Figure~\ref{fig:sample_scene} shows some sample frames from the DIVA dataset.

\begin{figure*}[h]
    \centering
    \includegraphics[width=1\linewidth]{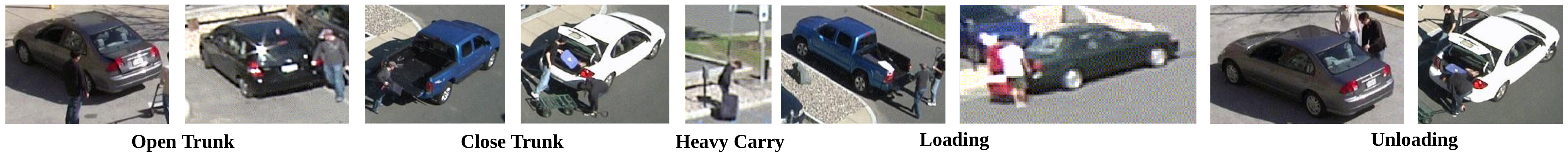}
    \caption{Sample frames of some activities of the DIVA dataset.} 
\label{fig:sample_activity}
\vspace{-5mm}
\end{figure*}

In this work we introduce a proposal-based modular system for performing spatio-temporal action detection in untrimmed videos. Our system generates spatio-temporal action proposals based on detections from an off-the-shelf detector, then classifies the proposals using an existing network architecture with minor changes.

Our proposed system has the advantage that it is both simple and does not require tracking of moving objects. Tracking of objects, often seen as an important component in action recognition systems, presents considerable challenges. Tracking errors generate problems in action recognition from which it is very difficult to recover. On the other hand, object detection has advanced consistently over the past few years, with more sophisticated frame-wise object detectors becoming available. These detectors can be successfully applied to previously unseen videos.

The proposed approach for action detection is based on the observation that we can generate high-recall proposals by clustering object detections. The dense proposals are then applied to a deep 3D-CNN to classify them as either one of the action classes or the non-action class. The temporal bounds for the proposals are also refined to improve localization. We modify the existing I3D~\cite{i3d17} network for action classification by adding an additional loss term for temporal refinement. We call the modified network Temporal Refinement I3D (TRI-3D). In summary, this paper makes the following contributions:
\vspace{-2.2mm}
\begin{itemize}
\setlength\itemsep{-1mm}
\item We introduce a proposal-based modular system for spatio-temporal action detection in untrimmed security videos.
\item We propose an algorithm using hierarchical clustering and temporal jittering for generating action proposals using frame-wise object detections.
\item We propose the Temporal Refinement I3D (TRI-3D) network for action classification and temporal localization.
\item We evaluate our system on the DIVA dataset, which is an untrimmed security video dataset in the wild.
\end{itemize}
\vspace{-1mm}


\vspace{-2mm}
\section{Related Work}
\vspace{-2mm}
\label{sec:related}

Much research has gone into designing algorithms for action classification from videos. In recent years, many methods have achieved remarkable performance using CNNs for the task of action classification~\cite{i3d14, karpathy2014large, negin2016human, yue2015beyond}. While~\cite{karpathy2014large, yue2015beyond} use frame-based features, \cite{i3d14} uses a two-stream (RGB and optical-flow) CNN approach to utilize the temporal information of videos. More recently, researchers have used 3D-CNNs for action classification~\cite{i3d17,tran2015learning,hou2017tube,saha2017amtnet} that simultaneously take in multiple video frames and classify them into actions.

 The task of spatio-temporal action detection from untrimmed videos is a challenging problem. Less work has gone into localizing actions, not just along the temporal axis but also in terms of spatial localization. Existing action detection algorithms can be broadly classified into 1) end-to-end systems, and 2) proposal-based systems. The end-to-end action detection approaches feed a chunk of video frames into a CNN which simultaneously classifies  and localizes the action.  Hou~et~al.~\cite{hou2017tube} proposed a tube convolutional neural network (T-CNN) that generates tube proposals from video-clips and performs action classification and localization using an end-to-end 3D-CNN. Kalogeiton~et~al.~\cite{kalogeiton2017action} extracts convolutional features from each frame of a video, and stacks them to learn spatial locations and action scores. Although these end-to-end learning methods may have a simpler pipeline, they are less effective for security videos, where the action is likely to happen in a small spatial region of a frame. On the other hand, proposal-based methods perform action detection in two steps. The first step computes the action proposals, while the second step classifies and localizes the action. Some proposal-based methods are presented in ~\cite{oneata2014spatio, zhu2017tornado, mettes2016spot, gkioxari2015finding, marian2015unsupervised}. Unlike existing proposal-based approaches, our method uses hierarchical clustering and temporal jittering to group frame-wise object detections obtained from off-the-shelf detectors in the spatio-temporal domain. It gives us the advantage of detecting variable length action sequences spanning a small spacial region of the video.




In this section we have covered a number of action recognition works, however this list is far from complete. For further action recognition works we point the reader to a more extensive curated list presented in ~\cite{awesomelist}.
\vspace{-2mm}
\section{DIVA dataset}
\begin{figure}
	\vspace{-2mm}
    \centering
    \includegraphics[width=1\linewidth]{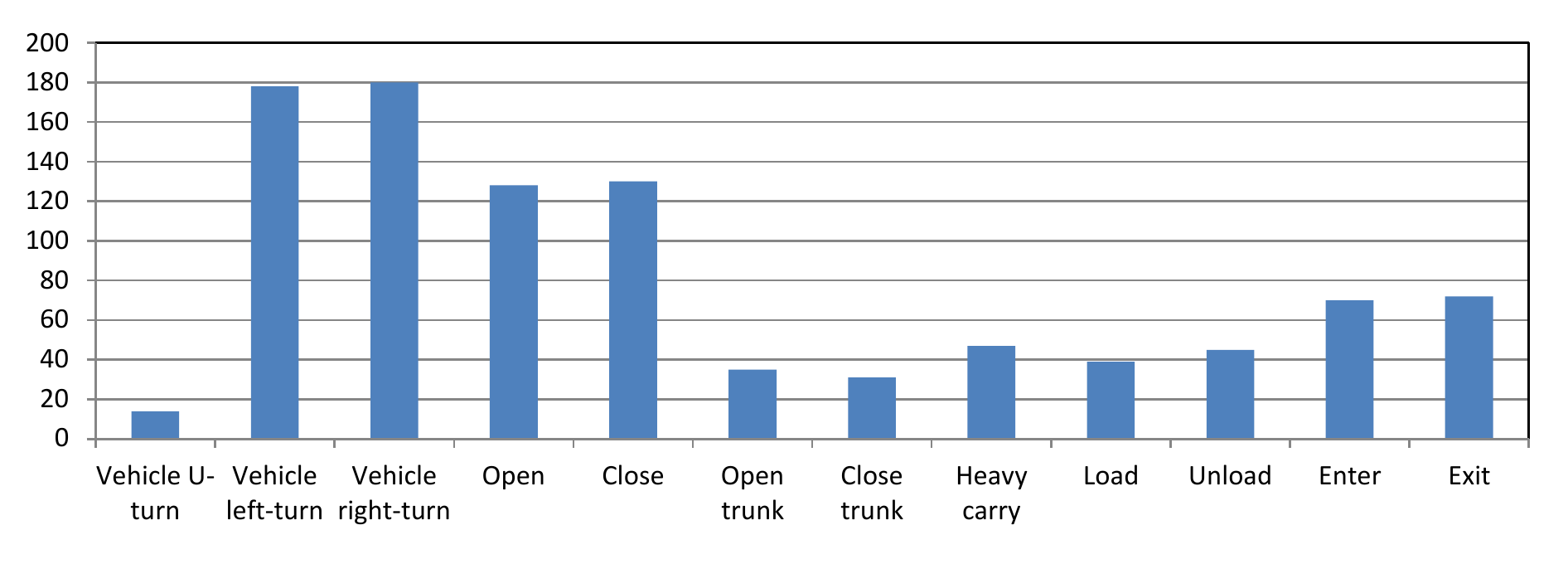}
    \caption{The number of training instances per action from the DIVA training set.} 
\label{fig:activity_n_insts}
\vspace{-5mm}
\end{figure}

\vspace{-2mm}
The DIVA dataset is a new spatio-temporal action detection dataset for untrimmed videos. While we present our work on the DIVA dataset, and there are currently no other papers we can cite that reference the DIVA dataset at this time, we would like to emphasize that we did not create the DIVA dataset, nor are we the only ones who have access to it. We would like to point out that a workshop on DIVA results will be organized as part of WACV 2019. The current release of the DIVA dataset (DIVA V1) is adapted from the VIRAT dataset~\cite{virat} with new annotations for 12 simple and complex actions of interest focusing on the public security domain. All actions involve either people or vehicles. Actions include \texttt{vehicle U-turn}, \texttt{vehicle left-turn}, \texttt{vehicle right turn}, \texttt{closing trunk}, \texttt{opening trunk}, \texttt{loading}, \texttt{unloading}, \texttt{transport heavy carry}, \texttt{open} (door), \texttt{close} (door), \texttt{enter}, and \texttt{exit}. The dataset currently consists of 455 video clips with 12 hours and 40 minutes in total captured at different sites. There are 64 videos in the training set, 54 videos in the validation set, 96 videos with annotations withheld in the test set. The remaining videos are for future versions of the dataset. All video resolutions are either $1200 \times 720$ or $1920 \times 1080$ and humans range in height from $20$ to $180$ pixels. We show sample images of different scenes and sample frames of some activities in Figure~\ref{fig:sample_scene} and  Figure~\ref{fig:sample_activity} respectively. In addition, the number of training instances per action is shown in Figure~\ref{fig:activity_n_insts}.

\vspace{-2mm}
\subsection{Challenges}
\vspace{-2mm}

As compared to other action detection datasets, such as the THUMOS'14~\cite{thumos} and AVA~\cite{ava} datasets, the DIVA dataset introduces several new challenges for the action detection task that make methods designed for existing action datasets unsuitable. The first issue is the sparsity of actions, both spatially and temporally. For example, exactly half of all videos contain at least 30 seconds of footage where no actions are performed. What makes DIVA particularly challenging is the spatial sparsity of actions: the average size for the bounding boxes of all actions in the training set is $264 \times 142$. As a result, when an action is occurring it only takes up on average less than $2.6\%$ of the pixels in any given image, and no action in the entire dataset takes up more than $40\%$. Additionally, with few exceptions, the similarity of each action and each environment makes it very difficult to use the context of the surrounding scene to assist in classification.
%
%

When compared with other setups, where actions are assumed to make up the majority of pixels on any given frame,  this motivates the need for a completely different approach focused on the localization of activities. For example, \cite{ren2015faster} mention that the smallest anchor size in Faster-RCNN is $128 \times 128$ on a $600 \times 600$ input image, or $4.5\%$ of the image pixels. This means the average action in DIVA is barely more than half the size of the smallest object detectable by conventional means.

The dataset also contains significant spatial and temporal overlap between activities. This is not just an issue between unrelated activities in the same frame (\eg one person entering a car while another leaves a different car), but is actually more fundamental. For a common example, consider the activities  \texttt{opening},  \texttt{entering}, and  \texttt{closing}, which apply to a human actor interacting with a car.  In order to enter a car, a subject may first open the car door (though it may already be open), and will often close it afterwards (though they may not). All three of these actions are usually performed in quick succession, yet DIVA begins annotation of each activity one second before it begins, and finishes annotating one second after it completes. It is therefore imperative that our system can handle large degrees of spatio-temporal overlap.

\begin{figure}
    \centering
    \includegraphics[width=1\linewidth]{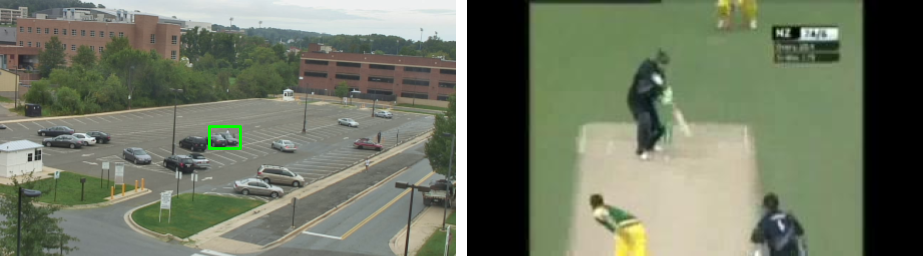}
    \caption{On the left, the DIVA action \texttt{Closing} makes up only a small portion of the image, and the surrounding context has no value for the action classification task. The THUMOS action \texttt{Cricket} on the right is much larger in the image, and the entire image's context is useful for classification.} 
\label{fig:diva_v_thumos}
\vspace{-2mm}
\end{figure}

\vspace{-2mm}
\section{Proposed Approach}\label{sec:method}
\vspace{-2mm}

\begin{figure*}
\centering
\includegraphics[width=0.9\textwidth]{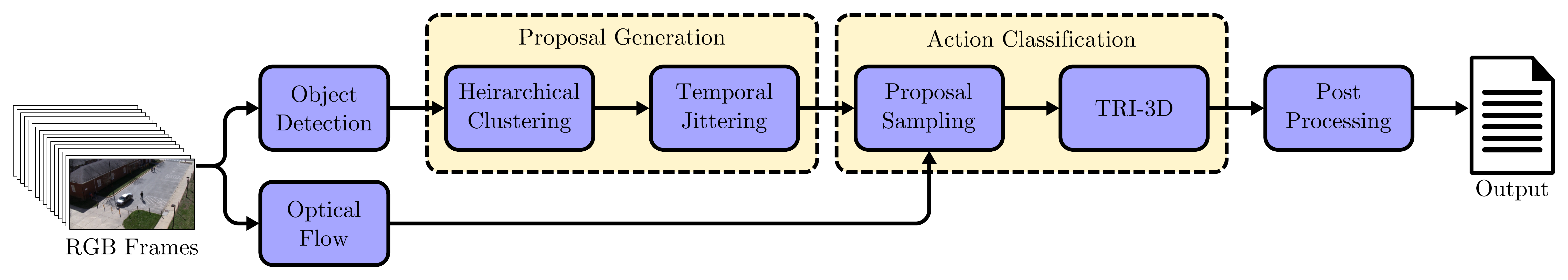}
\caption{Proposed system for spatio-temporal action detection.}
\label{fig:system}
\vspace{-5mm}
\end{figure*}

Our approach consists of three distinct modules. The first one generates class-independent spatio-temporal proposals from a given untrimmed video sequence. The second module performs action classification and temporal localization on these generated proposals using a deep 3D-CNN. The final module is a post-processing step that performs 3D non-maximum suppression (NMS) for precise action detection. The system diagram for our approach is shown in Figure~\ref{fig:system}. In the following sub-sections we discuss in detail the steps of our proposed approach.

\vspace{-2mm}
\subsection{Action Proposal Generation}
\vspace{-2mm}

The primary goal of the action proposal generation stage is to produce spatio-temporal cuboids from a video with high recall and little regard for precision. Although sliding-window search in spatio-temporal space is a viable method for proposal generation, it is computationally very expensive. An alternate solution is to use unsupervised methods to cluster the spatio-temporal regions from a video in a meaningful way. In our approach, we generate the action proposals by grouping frame-wise object detections obtained from Mask-RCNN~\cite{he2017mask} in the spatio-temporal domain using hierarchical clustering. These generated proposals are further jittered temporally to increase the overall recall.

\begin{figure}
    \centering
    \includegraphics[width=1\linewidth]{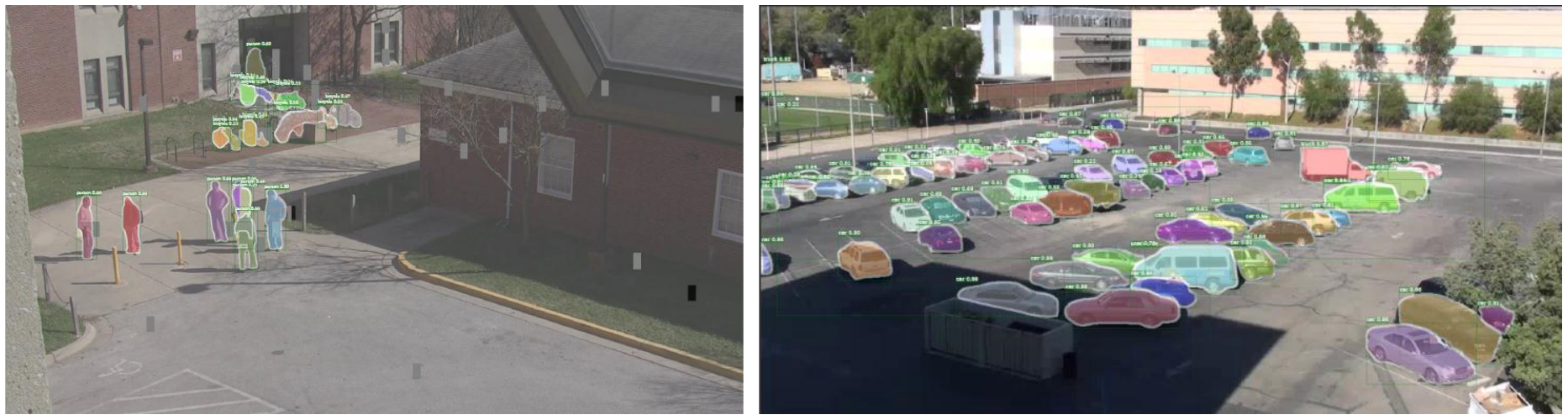}
    \caption{Some of object detection and segmentation results for the DIVA dataset using Mask-R-CNN~\cite{he2017mask}.} 
\label{fig:detectron}
\vspace{-5mm}
\end{figure}

\vspace{-2mm}
\subsubsection{Object Detection}
\vspace{-2mm}

For object detection, we apply Mask R-CNN~\cite{he2017mask}, an extension of the well-known Faster R-CNN~\cite{ren2015faster} framework. In addition to the original classification and bounding box regression network of Faster R-CNN, Mask-RCNN adds another branch to predict segmentation masks for each Region of Interest (RoI).
In Figure~\ref{fig:detectron}, we show some sample results from video frames of the DIVA dataset. We observe that Mask-RCNN is able to detect humans and vehicles at different scales, a feature which is useful for detecting the multi-scale actions of the DIVA datasets.

\vspace{-2mm}
\subsubsection{Hierarchical Clustering}
\vspace{-2mm}

The objects detected using Mask-RCNN are represented by a 3-dimensional feature vector $(\mathbf{x}, \mathbf{y}, \mathbf{f})$, where $(\mathbf{x}, \mathbf{y})$ denotes the center of the object bounding box and $\mathbf{f}$ denotes the frame number. We use the SciPy implementation of Divisive Hierarchical Clustering \cite{clustering, scipy} to generate clusters from the 3-dimensional features. We dynamically split the resulting linkage tree at various levels to create $k$ clusters, where $k$ is proportional of video length. The proposals are generated from the bounding box of all detections in the cluster. They are cuboids in space-time and are denoted by $(x_{min}, y_{min}, x_{max}, y_{max}, f_{start}, f_{end})$. This yields an average of approximately $250$ action proposals per video on DIVA validation set. Further details regarding the exact implementation can be found in the supplementary material.

\vspace{-2mm}
\subsubsection{Dense Action Proposals with Temporal Jittering}
\vspace{-2mm}

Although the proposals generated using hierarchical clustering reduce the spatio-temporal search space for action detection, they are unable to generate high recall for the following two reasons: 1) The generated proposals are independent of both the action class and cuboid temporal bounds. Hence, they are less likely to overlap precisely with the ground-truth action bounds. 2) Few proposals are generated. A higher recall is achieved with larger numbers of proposals. In order to solve these issues, we propose a temporal jittering approach to generate dense action proposals from the existing proposals obtained from hierarchical clustering. 

Let the start and end frames for an existing proposal be denoted by $f_{st}$ and $f_{end}$ respectively. We first choose the anchor frames by sliding along the temporal axis from $f_{st}$ to $f_{end}$ with a stride of $\textit{s}$.
\begin{figure}[h]
\vspace{-3mm}
    \centering
     \includegraphics[width=1\linewidth]{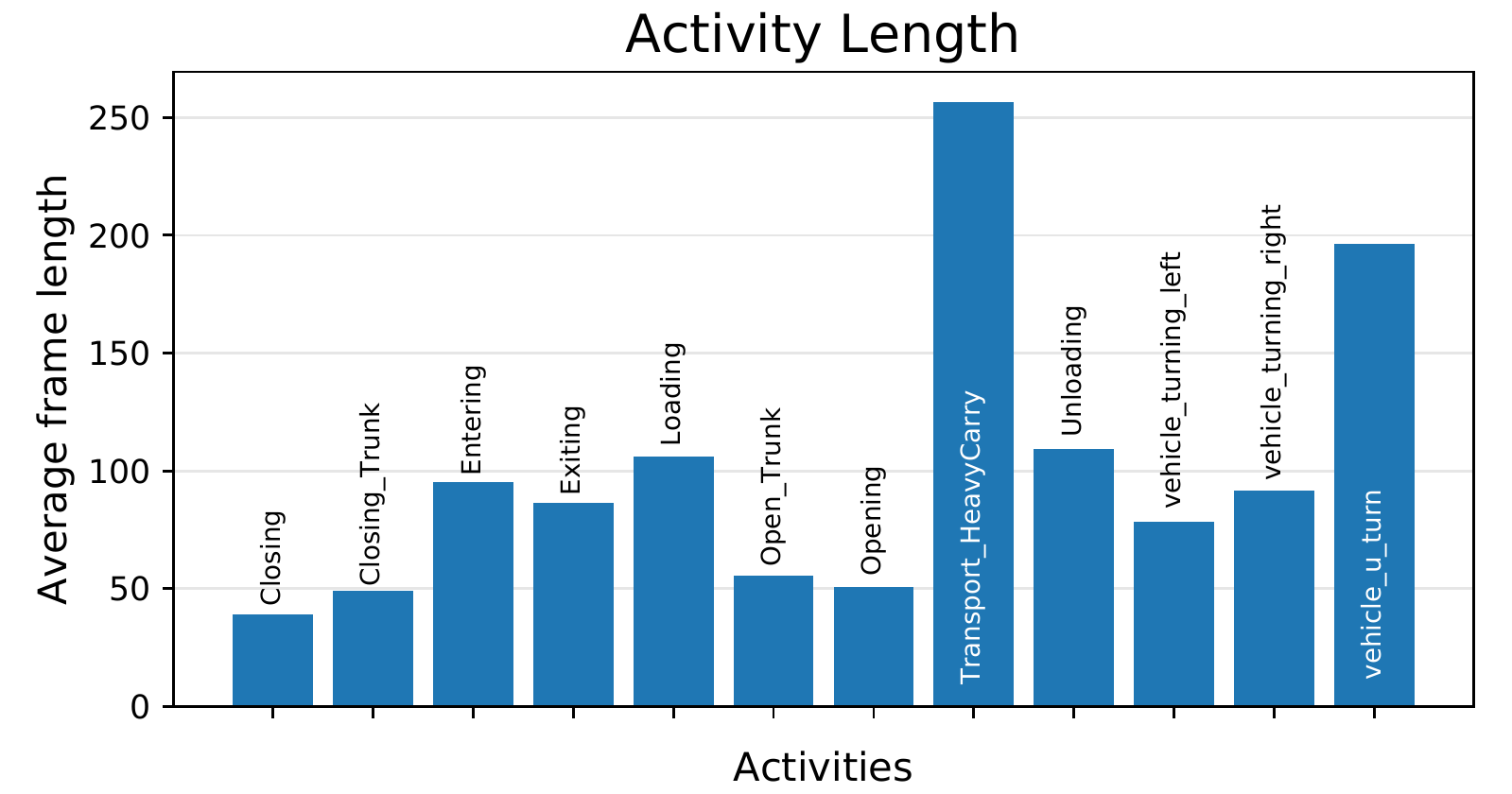}
    \caption{The average frame lengths for $12$ different activities from the DIVA training set.} 
\label{fig:activity_lengths}
\vspace{-2mm}
\end{figure}
The anchor frames thus selected are $(f_{st}, f_{st}+s, f_{st}+2s, f_{st}+3s, ..., f_{end})$. For each of the anchor frames $f_{a}$, we generate four sets of proposals with temporal bounds $(f_{a}-16, f_{a}+16)$, $(f_{a}-32, f_{a}+32)$, $(f_{a}-64, f_{a}+64)$ and $(f_{a}-128, f_{a}+128)$. We choose the proposals frame lengths to be $\{32, 64, 128, 256\}$ based on the average frame lengths for the actions in the DIVA dataset which range between $32-256$~(see Figure~\ref{fig:activity_lengths}). The pseudo-code for generating dense action proposals is presented in Algorithm~\ref{alg:tempjitt}.

\vspace{-2mm}
\begin{algorithm}
\small
\caption{Dense Proposal Generation}\label{alg:tempjitt}
\begin{algorithmic}[1]

\State $\textbf{detections}\gets Mask-RCNN (\textbf{video})$
\State $\textbf{orig\_proposals}\gets hierarchical\_clustering (\textbf{detections})$
\State $\textbf{new\_proposals}\gets \textbf{orig\_proposals}$
\State $\textbf{s}\gets 15$
\For{\texttt{proposal \textbf{in} orig\_proposals}}
	\State $\mathbf{x_{0}}, \mathbf{y_{0}}, \mathbf{x_{1}}, \mathbf{y_{1}} \gets spatial\_bounds(\textbf{proposal})$
    \State $\mathbf{f_{st}}, \mathbf{f_{end}} \gets temporal\_bounds(\textbf{proposal})$
    \For{\texttt{f from $\mathbf{f_{st}}$ to $\mathbf{f_{end}}$ step $\mathbf{s}$}}
    \State $\textbf{new\_proposals}.add(\mathbf{f}-16, \mathbf{f}+16)$
    \State $\textbf{new\_proposals}.add(\mathbf{f}-32, \mathbf{f}+32)$
    \State $\textbf{new\_proposals}.add(\mathbf{f}-64, \mathbf{f}+64)$
    \State $\textbf{new\_proposals}.add(\mathbf{f}-128, \mathbf{f}+128)$
    \EndFor
\EndFor
\State $\textbf{final\_dense\_proposals}\gets \textbf{new\_proposals}$
\end{algorithmic}
\end{algorithm}
\vspace{-2mm}

Generating dense proposals using temporal jittering has two advantages. First, the recall is higher. Second, having a large number of dense training proposals provides better data augmentation for training, thus improving performance.

Figure~\ref{fig:proposal_recall} provides a quantitative comparison of three action proposal generation methods on the DIVA validation set in terms of recall vs 3D Intersection over Union (IoU) with the ground-truth activities: 
\begin{figure}[h]
    \centering
    \includegraphics[width=1\linewidth]{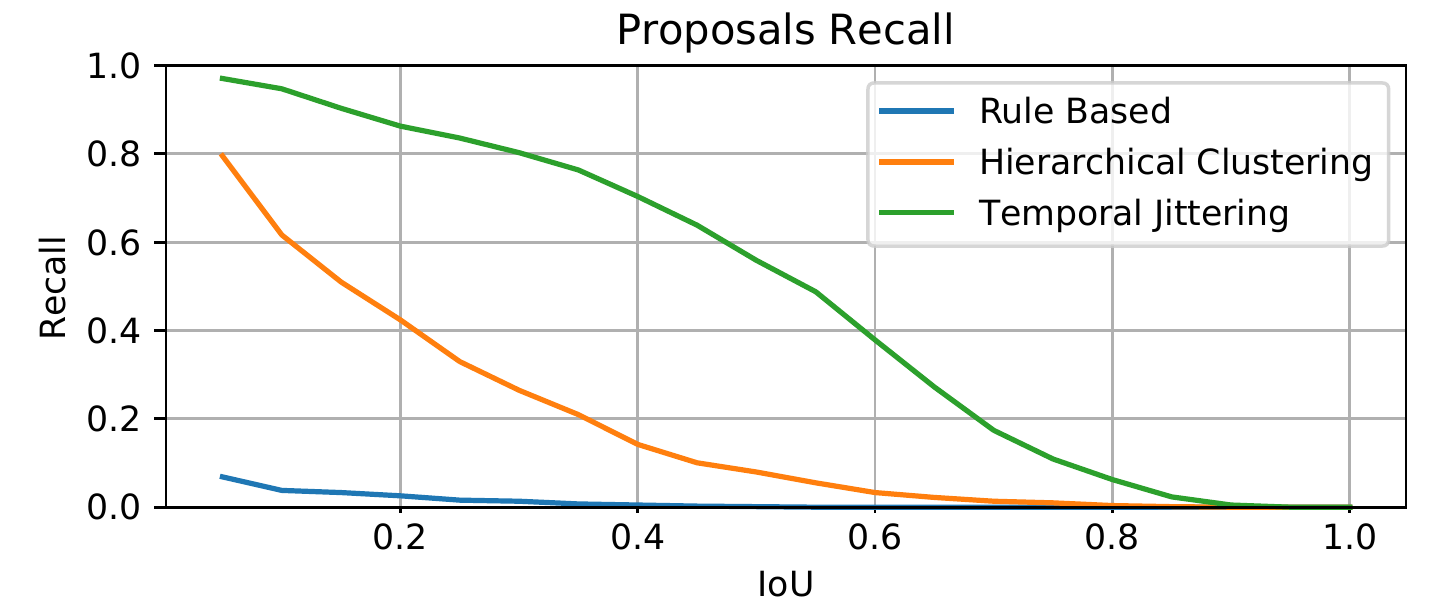}
    \caption{The recall for different action-proposal generation methods as a function of spatio-temporal intersection over union (IoU) overlap on the DIVA validation set.} 
\label{fig:proposal_recall}
\end{figure}
1) Rule based 2) Hierarchical clustering, and 3) Temporal Jittering.
The rule based proposal generation method uses hand-crafted rules to associate detections across consecutive frames. For example, a rule stating ``a person detection and a car detection closer than $50$ pixels is a proposal'' can be used to generate action proposals using Prolog. However, the recall with our rule based proposal generation method is poor. On the other hand, hierarchical clustering provides $40\%$ recall at a spatio-temporal IoU of $0.2$. Using temporal jittering on top of it increases the recall to $85\%$. This shows the effectiveness of our dense proposal generation method.

\vspace{-1mm}
\subsection{Action Proposal Refinement and Classification}
\vspace{-2mm}

The proposal refinement and classification step takes in the generated dense proposals as input and performs the following tasks:

\vspace{-3mm}
\begin{itemize}
\setlength\itemsep{-1mm}
\item Identify the non-action proposals.
\item Identify the action proposals and classify them into one of the given action classes.
\item Improve the temporal localization of the action proposals by refining their temporal bounds.
\end{itemize}
\vspace{-3mm}

\begin{figure}
  \centering
  \includegraphics[width=0.45\textwidth]{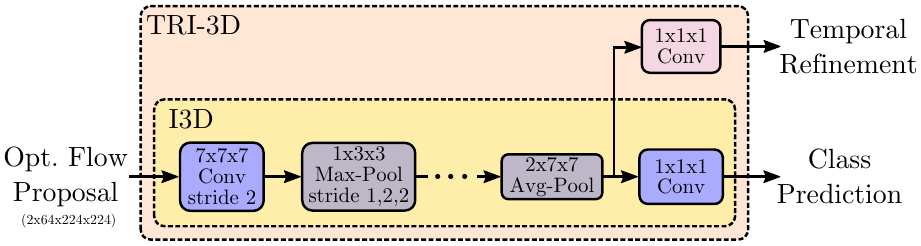}
  \caption{TRI-3D network. For details on the I3D please refer to Figure 3 in~\cite{i3d17}.}
  \label{fig:tri3d}
  \vspace{-5mm}
\end{figure}

\noindent To accomplish these tasks we begin with the I3D network architecture~\cite{i3d16} which classifies each proposal cuboid as one of the 12 action classes or as a non-action class. Due to the sparsity of action proposals, many of the ground truth actions don't perfectly correspond to any of the action proposals. To mitigate this issue, a regression objective is added to the final layer of I3D to predict a temporal correction to the cuboid. This defines the TRI-3D network. Figure~\ref{fig:tri3d} depicts the network architecture of the proposed TRI-3D. A detailed description of input pre-processing and the working of TRI-3D is provided in the following sub-sections.

\vspace{-2mm}
\subsubsection{Input Pre-processing}
\vspace{-2mm}

The input to the TRI-3D network is $64 \times 224 \times 224$ optical flow frames which are computed using the TV-L1 optical flow algorithm~\cite{tvl1}. The original I3D is designed for full sized frames on the Kinetics dataset. It first scales the input videos so that the smallest dimension is equal to $256$, then samples a random $224 \times 224$ crop during training. Unlike the Kinetics dataset, the aspect ratio of generated dense proposal cuboids is arbitrary. To address this issue we pad the smaller side of the proposal relative to the center so that it is the same size as the largest side prior to sampling the optical flow data. This has the potential to add a significant amount of extra padding to proposal cuboids that have a significant discrepancy between the dimensions; however, more conservative padding and cropping schemes performed poorly during preliminary testing (Section \ref{sec:ablation}).

\begin{figure}[h]
  \vspace{-4mm}
  \centering
  \includegraphics[width=0.35\textwidth]{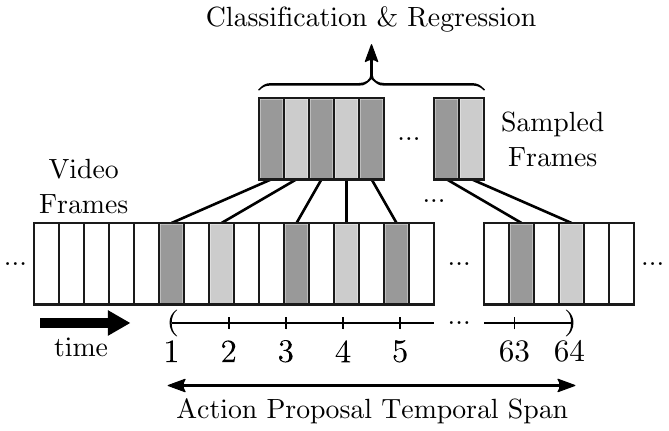}
  \caption{Uniform sampling 64 input frames from an arbitrarily long action proposal.}
  \label{fig:unifsample}
  \vspace{-5mm}
\end{figure}

A proposal may also extend across an arbitrary number of frames. However, since I3D requires a fixed number of input frames, we chose to uniformly sample the fixed number of frames across the temporal span of each proposal as shown in Figure~\ref{fig:unifsample}. This strategy follows the original I3D work in many cases. The sampling strategy differs from the original work when the total number of frames in a proposal is less than $64$. In this scenario, instead of wrapping the sampling we continue to sample uniformly, potentially sampling the same frame multiple times in a row. This strategy ensures frames are always provided to the network in temporally increasing order which is necessary for temporal refinement.

After sampling the cropped cuboid from the optical flow video, the sample is then resized to $256 \times 256$. During training we randomly sample a $224 \times 224$ crop from the cuboid.


To improve the robustness of the network we apply random horizontal flips on all training examples except the non-symmetric actions: \texttt{vehicle turning left}, \texttt{vehicle turning right}, and \texttt{vehicle u turn}.

The TRI-3D network uses optical flow frames as input. The two-stream I3D network has been shown to outperform the single stream network on the Kinetics dataset, however through experimentation we discovered that the two-stream network performed worse for overall prediction accuracy when using the ground truth actions, which motivated our choice to use only optical flow, see Section~\ref{sec:ablation}. An additional benefit of this choice is an improvement in classification speed.

Since the videos in the dataset are from stationary cameras we don't perform any normalizations to the optical flow to adjust the mean value.

\vspace{-2mm}
\subsubsection{TRI-3D Training}
\vspace{-2mm}

The target data for the TRI-3D network is derived from proposals generated from training data and labeled from ground truth annotations. We label each proposal produced by the proposal generator as either one of the action classes or the non-action class. If a proposal is found to have a sufficient IoU overlap with a ground truth annotation, then a temporal refinement value is also calculated. A proposal with spatial IoU above $0.35$ and temporal IoU above $0.5$ with the ground-truth is designated as a ``positive'' proposal and assigned an action class label. A proposal with temporal IoU less than $0.2$ is designated as a ``negative'' proposal. Negative proposals are further separated into two designations, easy negatives and hard negatives. Any negative proposal with spatial IoU above $0.35$ and temporal IoU between $0.01$ and $0.2$ with ground truth is a hard negative, and the remaining negatives are easy negatives.

In order to improve the robustness of our network with respect to false alarms we construct our training data using all of the proposals which were assigned a class label, as well as all of the hard negatives. A total of $9,525$ easy negatives are generated, however most of the easy negatives are omitted from the training data. The only easy negatives used are those which result from hierarchical clustering, omitting those derived from temporal jittering. The total number of training examples for each designation are shown in Table~\ref{tbl:training}


\begin{table}[h]
  \vspace{-2.25mm}
  \centering
  \begin{tabular}{c c c | c}
  Positive & Easy Neg. & Hard Neg. & Total \\
  \cline{1-4}
  \textbf{12,752} & \textbf{9,525} & \textbf{13,574} & \textbf{35,851}
  \end{tabular}
  \caption{Numbers of proposals of each type used during training of our TRI-3D network.}
  \label{tbl:training}
  \vspace{-3.5mm}
\end{table}

The number of training samples generated for different action classes vary widely. For instance, the number of positive samples for \texttt{vehicle u turn} is $215$, while it's $2,554$ for \texttt{vehicle right turn}. To mitigate the effect of the high class imbalance, we duplicate the positive instances of each class such that all the action classes have an equal number of positive samples during training. This helps the network to learn equally discriminative features for all the action classes involved in training.

\vspace{-3mm}
\subsubsection{TRI-3D Loss Functions}
\vspace{-2mm}
The TRI-3D network takes in the preprocessed proposals and performs action classification as well as temporal refinement. For training action classification, we use a multi-class cross-entropy loss function $L_{cls}$ as shown in (\ref{eq:loss_cls})

\vspace{-2.5mm}
\begin{equation}
\label{eq:loss_cls}
L_{cls} = \sum \limits_{a=0}^{12} -y_{a} \cdot \log(p_{a}),
\end{equation}
\vspace{-2.5mm}

\noindent where $y_{a} = 1$ if the sample belongs to class $a$, otherwise $0$. The predicted probability that a sample belongs to class $a$ is given by $p_{a}$. The $12$ action classes are labeled from $a=1$ to $a=12$, while the non-action class is labeled $a=0$. 

We also refine the temporal bounds of the proposals by adjusting its start and end frames according to the the predicted regression values. Let the start and end frames for the input proposal and the ground-truth be ($f_{st}$ ,$f_{end}$) and ($\hat{f}_{st}$ ,$\hat{f}_{end}$), respectively. We select the mid-frame of the proposal as $f_{a} = (f_{st} + f_{end}) / 2$, and the half-length of the proposal as $t = (f_{end} - f_{st} + 1) / 2$. The normalized ground truth regression pair is generated by~(\ref{eq:temp_norm}):

\vspace{-3mm}
\begin{equation}
\label{eq:temp_norm}
(r_{st},r_{end}) = \left(\frac{\hat{f}_{st}-f_{a}}{t},\frac{\hat{f}_{end}-f_{a}}{t}\right).
\end{equation}
\vspace{-3mm}

\noindent We use the $\mathrm{smooth}_{L1}$ loss~\cite{ren2015faster} to generate the regression outputs $v = (v_{st},v_{end})$ for the ground-truth labels $r = (r_{st},r_{end})$, as shown in~(\ref{eq:smooth_l1_temp})
\vspace{-2.5mm}
\begin{align}
\begin{split}
\label{eq:smooth_l1_temp}
L_{loc} (v, r) = &~\mathrm{smooth}_{L1} (r_{st} - v_{st}) \\
				 &+ \mathrm{smooth}_{L1} (r_{end} - v_{end}),
\end{split}
\end{align}
\vspace{-3mm}
\noindent in which
\begin{equation}
\label{eq:smooth_l1}
\mathrm{smooth}_{L1}(x) = \begin{cases}
       0.5x^{2} &\quad\text{if } |x|<1 \\
       |x| - 0.5 &\quad\text{otherwise.} \\  
     \end{cases}
\end{equation}
\vspace{-3mm}

We combine the action classification and the temporal regression loss in a multi-task fashion as given by~(\ref{eq:full_loss}) 

\vspace{-3mm}
\begin{equation}
\label{eq:full_loss}
L_{full} = L_{cls} + \lambda [a \geq 1] L_{loc},
\end{equation}
\vspace{-5mm}

\noindent where $\lambda$ is chosen to be $0.25$. The temporal refinement loss is activated only for the positive proposals that belong to one of the $12$ action classes ($a \geq 1$). For a non-class proposal, the temporal refinement loss doesn't generate any gradients. We train the TRI-3D network using the Adam~\cite{kingma2014adam} optimization technique, with initial learning rate set to $0.0005$.







\subsection{Post-processing}
\vspace{-2mm}

At test time, the TRI-3D network outputs the classification score and the refined temporal bounds for an input proposal. Our method for proposal generation creates many highly-overlapping action proposals, many of which are classified as the same class. We prune overlapping cuboids using 3D-NMS. The 3D-NMS algorithm is applied to each of the classes separately and considers two proposals to be overlapping when the temporal IoU overlap is greater than $0.2$ and the spatial IoU overlap is greater than $0.05$.

\vspace{-1mm}
\section{Experimental Evaluation}\label{sec:exp}
\vspace{-2mm}
In this section we present and discuss various experiments which motivate our design choices and describe the overall system performance. All experimental results are reported on the DIVA validation dataset unless otherwise indicated. 





\subsection{DIVA Evaluation Metric}
\vspace{-2mm}

To correctly and objectively assess the performance of the proposed action detection (AD) framework on DIVA, we adopt the measure: probability of missed detection $P_{miss}$ at fixed rate of false alarm per minute $Rate_{FA}$, which is used in the surveillance event detection framework of TRECVID 2017~\cite{trecvid2017}. This metric evaluates whether the algorithm correctly detects the presence of the target action instances. A one-to-one correspondence from detection to ground-truth is enforced using the Hungarian algorithm, thus each detected action may be paired with either one or zero ground-truth actions. Any detected action which doesn't correspond to a ground-truth action is a false alarm, and any ground-truth action which isn't paired with a corresponding detection is a miss. The evaluation tool used in this work is available on Github~\cite{actevscorer}. For details of the evaluation metric we refer the reader to TRECVID 2017~$\cite{trecvid2017}$.





\subsection{Preliminary Classification Experiments} 
\vspace{-2mm}

\begin{figure*}
  \centering
  \includegraphics[width=0.9\textwidth]{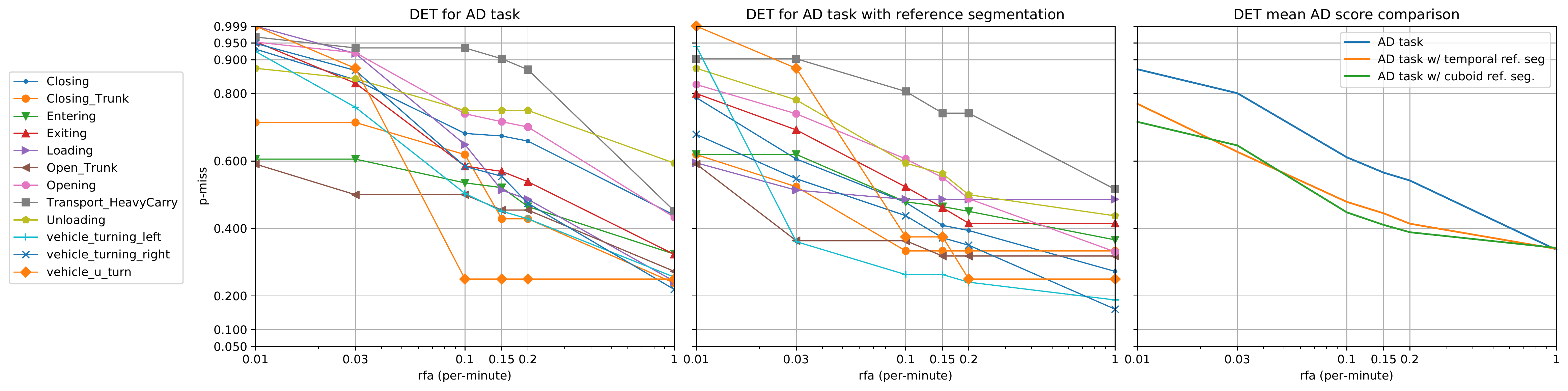}
  \caption{The left plot shows the per-class DET curves for the AD task. The center plot shows the per-class DET curves for TRS. The right plot shows the aggregated DET performance on the AD task compared to TRS and CRS.}
  \label{fig:ad_results}
  \vspace{-5mm}
\end{figure*}

\begin{table}
\centering
\small
\begin{tabular}{c c c c | c}
Net Arch. & Input Mode & Pretrained & Crop & Acc \\
\cline{1-5}
I3D & opt. flow & True & square & \textbf{0.716} \\
I3D & \textit{RGB} & True & square & 0.585 \\
I3D & \textit{RGB+flow} & True & square & 0.704 
\end{tabular}
\normalsize
\caption{Classification accuracy for the preliminary classifier study on ground truth proposals from the DIVA validation set. The first row represents our final design. Subsequent rows contain experimental results with various modifications.}
\label{tbl:ablation}
\vspace{-4mm}
\end{table}

\label{sec:ablation}
While developing our classification system we performed a number of preliminary experiments. The results of these experiments were used to justify the choices of network architecture, cropping scheme, and input modality. The experiments were performed by training on cuboids derived from ground truth annotations. Classification accuracy is reported on the DIVA validation ground truth cuboids.

The results of some of these experiments are shown in Table \ref{tbl:ablation}. One of the most surprising results was the discovery that the "single-stream" optical flow I3D network outperformed the two-stream I3D network. This is surprising because for other action recognition datasets two-stream networks generally outperform single stream networks \cite{i3d14, i3d16}. The output of the two-stream I3D network is computed by averaging the logits of the optical flow and RGB networks, which are trained independently. The table shows that the RGB I3D network is 13\% less accurate than the optical flow network. The poor performance of the two-stream network appears to be due to the poor performance of the RGB I3D network.

Further experiments to motivate our choice of architecture and cropping scheme were performed and can be found in the Supplementary Material.

\subsection{System Performance}
\vspace{-2mm}
In this section we present our experiments and results for the entire system. The primary goal of the system is to take untrimmed videos as input and report the frames where actions are taking place. We perform the following experiments in order to gain a better understanding of the system performance, to discover how much impact further improvements to proposal generation may have, and how much error is simply due to improper classification.
\vspace{-1pt}
\begin{itemize}
\setlength\itemsep{-1mm}
\item \textit{Action detection (AD)}: This is the primary task of the system. Given an untrimmed video, detect and classify the begin and end frame of each action.
\item \textit{AD with temporal reference segmentation (TRS)}: Perform the AD task but with additional temporal reference segmentation, that is, our system is provided the beginning and ending frames of each ground truth action, but not the class or spatial bounds.
\item \textit{AD with cuboid reference segmentation (CRS)}: Perform the AD task with both spatial and temporal reference segmentation. In this experiment the system is provided both the start and end frame of each activity as well as the spatial cuboid bounds for each activity.
\end{itemize}

In order to perform the TRS experiment we add an additional processing step between proposal generation and classification. We adjust the temporal bounds of any proposal which temporally overlaps a reference action. If a proposal overlaps multiple reference actions then multiple copies of the proposal are generated and the temporal bounds of the copies are adjusted to match each of the reference actions. Any proposals which have no temporal overlap with any reference actions are omitted. For this experiment we did not retrain the network and the temporal refinement values are ignored.

The CRS experiment is effectively performing action classification on reference cuboids. This is the same as the preliminary experiments described in Section \ref{sec:ablation}. The results described here correspond to the experiment referred to in the first row of Table \ref{tbl:ablation}.


The results of the experiments are shown in Figure \ref{fig:ad_results}. From these figures we see that there is a high degree of performance variation among different classes. For the AD task the worst performing class at $Rate_{FA}$ of $0.1$ is \texttt{Transport Heavy Carry} with $P_{miss}$ of $0.935$ and the best performing class is \texttt{vehicle u-turn} with a $P_{miss}$ of $0.25$. However, we see that at $Rate_{FA}$ of $0.01$ the \texttt{vehicle u-turn} becomes the worst performing action. This appears to be due to the fact that very few training and validation examples are available.


The aggregated results described in Figure \ref{fig:ad_results}, and equivalently in Table~\ref{tbl:ad_results}, show the significant improvement gained by providing temporal reference segmentation. Surprisingly, only a small average improvement is observed when providing the cuboid reference segmentation.

\textbf{DIVA Test Data}: The DIVA test dataset annotations are unavailable to the authors at the time of writing this manuscript. We submitted a single set of results from our system for independent evaluation and have included these results as well as results from other performers on the DIVA dataset in Table~\ref{tbl:test_diva}. We have also included results from a publicly available implementation based on~\cite{xu2017r}. We keep the identities of other performers anonymous.

\begin{table}[h]
\footnotesize
\centering
\begin{tabular}{c | c c c c c c}
$Rate_{FA}$ & 0.01 & 0.03 & 0.1 & 0.15 & 0.2 & 1.0 \\
\cline{1-7}
AD & 0.870 & 0.800 & 0.610 & 0.563 & 0.542 & 0.361 \\
Temporal Ref. & 0.770 & 0.627 & 0.479 & 0.445 & 0.414 & 0.340 \\
Cuboid Ref. & 0.716 & 0.646 & 0.448 & 0.410 & 0.389 & 0.342 \\
\end{tabular}
\vspace{-1mm}
\caption{Our system's mean $P_{miss}$ at fixed $Rate_{FA}$ on DIVA validation data. These are the values represented in the right plot in Figure \ref{fig:ad_results}.}
\label{tbl:ad_results}
\vspace{-1mm}
\end{table}

\begin{table}[h]
\footnotesize
\centering
\begin{tabular}{c | c c c c c c}
$R_{FA}$ & Xu \etal~\cite{xu2017r} & P4 & P3 & P2 & P1 & Ours \\
\cline{1-7}
0.15 & 0.863 & 0.872 & 0.759 & 0.624 & 0.710 & \textbf{0.618} \\
1.0 & 0.720 & 0.704 & 0.624 & 0.621 & 0.603 & \textbf{0.441}\\
\end{tabular}
\caption{Mean $P_{miss}$ versus $Rate_{FA}$ on the DIVA test data for AD task obtained via independent evaluation. DIVA performers (P1-P4) and algorithms other than the baseline~\cite{xu2017r} have been kept anonymous by request of the independent evaluator. Only the performers better than the baseline are represented here (sorted by mean $P_{miss}$ at $Rate_{FA}$ of $1$). A lower $P_{miss}$ value indicates superior performance, and the best performance is indicated in bold.}
\label{tbl:test_diva}
\vspace{-3mm}
\end{table}

\textbf{THUMOS'14 Dataset}: In order to compare to other action detection systems we also evaluated on the Temporal Action Detection task of THUMOS'14~\cite{thumos}. The THUMOS'14 Temporal Action Detection  dataset contains 200 annotated validation videos, and 213 annotated test videos for 20 action classes. Since the THUMOS'14 dataset is fundamentally different from DIVA in that actions generally span the majority of the frame, we omit the hierarchical clustering phase and instead perform temporal jittering on the cuboid spanning the entire video. 

Following common practice, we trained THUMOS'14 using the validation data for 5 epochs and then evaluated on the test data. The results are shown in Table~\ref{tbl:thumos}. From this table we see that our system, while not designed for videos of this nature, performs well compared to state-of-the-art methods.
\vspace{-1mm}


\begin{table}[h]
\footnotesize
\centering
\setlength\tabcolsep{4pt}
\begin{tabular}{c | c c c c c c c}
tIoU & 0.1 & 0.2 & 0.3 & 0.4 & 0.5 & 0.6 & 0.7 \\
\cline{1-8}
Karaman \etal~\cite{karaman2014fast} & 4.6 & 3.4 & 2.4 & 1.4 & 0.9 & - & - \\
Oneata \etal~\cite{oneata2014lear} & 36.6 & 33.6 & 27.0 & 20.8 & 14.4 & - & - \\
Wang \etal~\cite{wang2014action} & 18.2 & 17.0 & 14.0 & 11.7 & 8.3 & - & - \\
Caba \etal~\cite{caba2016fast} & - & - & - & - & 13.5 & - & - \\
Richard \etal~\cite{richard2016temporal} & 39.7 & 35.7 & 30.0 & 23.2 & 15.2 & - & - \\
Shou \etal~\cite{shou2016temporal} & 47.7 & 43.5 & 36.3 & 28.7 & 19.0 & 10.3 & 5.3 \\
Yeung \etal~\cite{yeung2016end} & 48.9 & 44.0 & 36.0 & 26.4 & 17.1 & - & - \\
Yuan \etal~\cite{yuan2016temporal} & 51.4 & 42.6 & 33.6 & 26.1 & 18.8 & - & - \\
Escorcia \etal~\cite{escorcia2016daps} & - & - & - & - & 13.9 & - & - \\
Buch \etal~\cite{buch2017sst} & - & - & 37.8 & - & 23.0 & - & - \\
Shou \etal~\cite{shou2017cdc} & - & - & 40.1 & 29.4 & 23.3 & 13.1 & 7.9 \\
Yuan \etal~\cite{yuan2017temporal} & 51.0 & 45.2 & 36.5 & 27.8 & 17.8 & - & - \\
Buch \etal~\cite{buch2017end} & - & - & 45.7 & - & 29.2 & - & 9.6 \\
Gao \etal~\cite{gao2017cascaded} & 60.1 & 56.7 & 50.1 & 41.3 & 31.0 & 19.1 & 9.9 \\
Hou \etal~\cite{hou2017real} & 51.3 & - & 43.7 & - & 22.0 & - & - \\
Dai \etal~\cite{dai2017temporal} & - & - & - & 33.3 & 25.6 & 15.9 & 9.0 \\
Gao \etal~\cite{gao2017turn} & 54.0 & 50.9 & 44.1 & 34.9 & 25.6 & - & - \\
Xu \etal~\cite{xu2017r} & 54.5 & 51.5 & 44.8 & 35.6 & 28.9 & - & - \\
Zhao \etal~\cite{zhao2017temporal} & \textbf{60.3} & 56.2 & 50.6 & 40.8 & 29.1 & - & - \\
Huang \etal~\cite{huang2018sap} & - & - & - & - & 27.7 & - & - \\
Yang \etal~\cite{yang2018exploring} & - & - & 44.1 & 37.1 & 28.2 & 20.6 & 12.7 \\
Chao \etal~\cite{chao2018rethinking} & 59.8 & \textbf{57.1} & 53.2 & \textbf{48.5} & \textbf{42.8} & \textbf{33.8} & \textbf{20.8} \\
Nguyen \etal~\cite{nguyen2018weakly} & 52.0 & 44.7 & 35.5 & 25.8 & 16.9 & 9.9 & 4.3 \\
Alwassel \etal~\cite{alwassel2018action} & - & - & 51.8 & 42.4 & 30.8 & 20.2 & 11.1 \\
Gao \etal~\cite{gao2018ctap} & - & - & - & - & 29.9 & - & - \\
Lin \etal~\cite{lin2018bsn} & - & - & \textbf{53.5} & 45.0 & 36.9 & 28.4 & 20.0 \\
Shou \etal~\cite{shou2018autoloc} & - & - & 35.8 & 29.0 & 21.2 & 13.4 & 5.8 \\
\cline{1-8}
Ours & 52.1 & 51.4 & 49.7 & 46.1 & 37.4 & 26.2 & 15.2 \\
\end{tabular}
\vspace{1mm}
\caption{Comparison to THUMOS'14 performers on the mAP metric at various temporal IoUs. Missing entries indicate that results are not available. We note that Xu \etal\cite{xu2017r} is the same system used to compute the DIVA V1 baseline; see Table \ref{tbl:test_diva}. The best performance at each tIoU is indicated in bold.}
\label{tbl:thumos}
\end{table}

\vspace{-2mm}
\section{Conclusion}\label{sec:conclusion}
\vspace{-2mm}

In this work we introduced an action detection system capable of handling arbitrary length actions in untrimmed security video on the difficult DIVA dataset. The system presented in this work is easily adapted to the THUMOS dataset. This system also leaves room for improvement and the modular design allows for easy integration of such future improvements. Although the DIVA evaluation metrics also includes an action-object detection task we choose to omit evaluation on this metric since our system doesn't explicitly provide object level localization of activities. We leave such improvements and extensions to future work.
\vspace{-5pt}




\vfill
\subsection*{Acknowledgments}\label{sec:acknowledge}
\vspace{-1mm}
This research is based upon work supported by the Office of the Director of National Intelligence (ODNI), Intelligence Advanced Research Projects Activity (IARPA), via IARPA R\&D Contract No. D17PC00345. The views and conclusions contained herein are those of the authors and should not be interpreted as necessarily representing the official policies or endorsements, either expressed or implied, of ODNI, IARPA, or the U.S. Government. The U.S. Government is authorized to reproduce and distribute reprints for Governmental purposes notwithstanding any copyright annotation thereon.

\clearpage

{\small
\bibliographystyle{ieee}
\bibliography{egbib}



\end{document}